# DETECTION AND CLASSIFICATION OF FAULTS AIMED AT PREVENTIVE MAINTENANCE OF PV SYSTEMS

**Edgar Hernando Sepúlveda Oviedo (1,2), Louise Travé-Massuyès (1,3), Audine Subias (1), Marko Pavlov (2), Corinne Alonso (1)**

1. LAAS-CNRS, Université fédérale de Toulouse, CNRS, INSA, UPS, France
2. Feedgy Solar, France
3. ANITI, Université fédérale de Toulouse, France;

*Núcleo Temático: Energías limpias*

## Introduction

Diagnosis in PV systems aims to detect, locate and identify faults. Diagnosing these faults is vital to guarantee energy production and extend the useful life of PV power plants. In the literature, multiple machine learning approaches have been proposed for this purpose. However, few of these works have paid special attention to the detection of fine faults and the specialized process of extraction and selection of features for their classification. A fine fault is one whose characteristic signature is difficult to distinguish to that of a healthy panel. As a contribution to the detection of fine faults (especially of the snail trail type), this article proposes an innovative approach based on the Random Forest (RF) algorithm. This approach uses a complex feature extraction and selection method that improves the computational time of fault classification while maintaining high accuracy.

## Methods

The approach proposed in this article is composed of four phases: *i)* Acquisition of the electric current signal; *ii)* Feature extraction using Multiresolution Signal Decomposition and statistical features; *iii)* Feature selection and *iv)* Fault classification using Random Forest (RF) algorithm. In the first phase, the current signal is captured for each panel every minute. In the second, the current signal is decomposed using an iterative decomposition based on wavelets [Ray 2018]. Then a set of statistical characteristics is extracted [Kurukuru 2020] from the coefficients of the decomposed signal. The third stage uses the PCA (Principal Component Analysis) algorithm to reduce the dimensionality of the feature matrix built in the third stage. Finally, the reduced feature matrix is used as input to the RF algorithm that builds multiple decision trees during the training phase and generates the final class by majority voting [Zhang 2020, Gunes 2003]. The operation of the RF algorithm is represented in Figure 1.

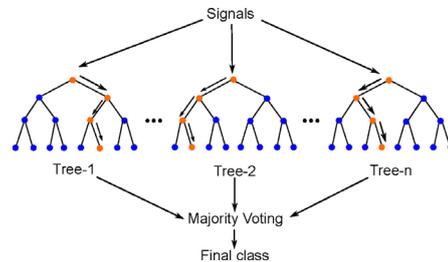

*Figure 1: Example of classifying a new signal using Random forest (RF).*

## Results

Figure 2 presents the current signals of the healthy panels and the snail trail panels during one day. As it can be seen in Figure 2, there is a high similarity between the two types of signals, which strongly hinders their correct classification. The classification results of the RF algorithm were evaluated using the $F_{score}$ and the confusion matrix. The $F_{score}$ of the RF is 0.80. $F_{score}$ varies between 0 and 1, with 0 being the worst value and 1 being a correct classification.

The classification results are presented in Figure 3, using the confusion matrix. As shown in Figure 3, RF mana ged to classify all the healthy panels and almost all (75%) of the panels with snail trail even with the reported high similarity between the signals of the two classes.

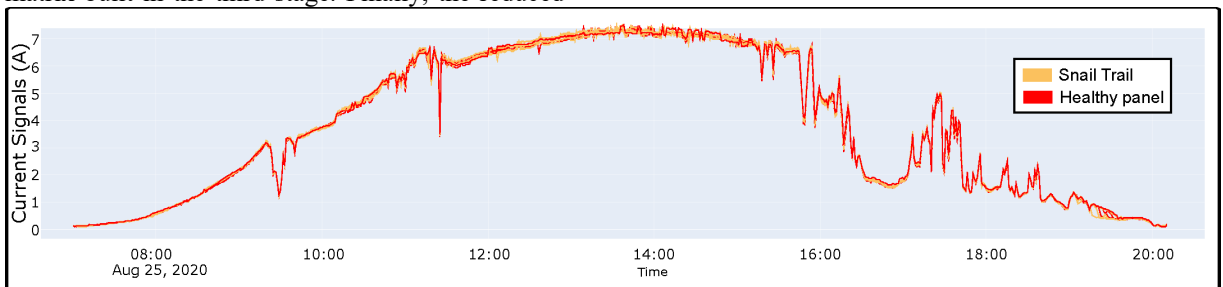

*Figure 2: 8 panel current signals (4 healthy and 4 with snail trail).*

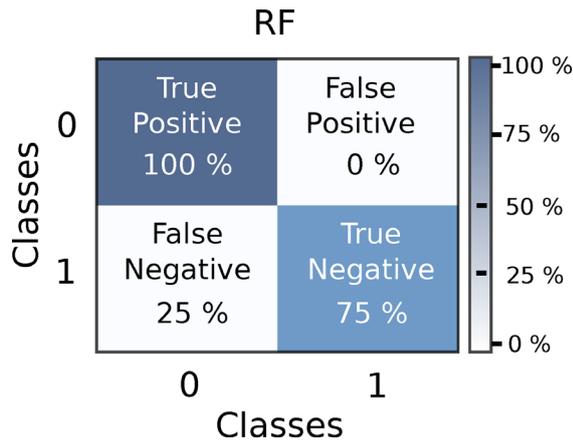

*Figure 3: Confusion matrix of the results of the RF model. Class 0 corresponds to healthy panels and class 1 corresponds to panels with a snail trail.*

## Discussion

The approach proposed in this article starts with $n$ randomly selected samples for training each tree in the RF algorithm. This process is done to increase the chances of training each tree with different samples.

RF combines several decision trees, instead of using a single learning model. This greatly increases the accuracy of fine fault detection. The approach presented in this article does not require multiple sensors and performs efficiently on snail trail fault classification just by capturing the current signal from the panels. Likewise, the proposed approach is capable of classifying the panels, both healthy and snail trail, despite a small number of training samples.